\pdfoutput=1
\documentclass[11pt]{article}

\usepackage{EMNLP2022}

\usepackage{times}
\usepackage{latexsym}
\usepackage{listings}
\usepackage[T1]{fontenc}
\usepackage[utf8]{inputenc}

\usepackage{graphicx}
\usepackage{ulem}
\usepackage{url}[hyphens]
\usepackage{amsmath}
\usepackage{breqn}
\usepackage{makecell}
\usepackage{tabularx}
\usepackage{booktabs}
\usepackage{multirow}
\usepackage[colorinlistoftodos]{todonotes}
\usepackage{textcomp}
\usepackage{comment}
\usepackage{bm}
\usepackage{hyperref}
\usepackage{amsmath}
\usepackage{multirow}
\usepackage{listings}
\usepackage[linesnumbered,ruled,vlined]{algorithm2e}
\usepackage{stfloats}
\usepackage{array}

\usepackage{microtype}

\usepackage[utf8]{inputenc}
\usepackage{microtype}
\usepackage{inconsolata}
\title{\textsc{Struc-Bench}: Are Large Language Models Good at Generating Complex Structured Tabular Data?}

\author{
\textbf{Xiangru Tang}$^\spadesuit$  \ \ 
        \textbf{Yiming Zong}$^{\heartsuit}$ \ \ 
         \textbf{Jason Phang}$^{\diamondsuit}$ \ \ 
                \textbf{Yilun Zhao}$^{\spadesuit}$ \ \ 
                  \textbf{Wangchunshu Zhou}$^{\spadesuit}$ \ \  \\
                 \textbf{Arman Cohan}$^{\spadesuit}$ \quad 
        \textbf{Mark Gerstein}$^\spadesuit$ \ \ 
       \\   
  $^\spadesuit$ Yale
University  \quad
  $^\heartsuit$ Zhejiang University  \quad
  $^{\diamondsuit}$ New York University \quad
  \\
  {\tt xiangru.tang@yale.edu}
}
\begin{document}
\maketitle
\begin{abstract}

\end{abstract}
Despite the remarkable capabilities of Large Language Models (LLMs) like GPT-4, producing complex, structured tabular data remains challenging. Our study assesses LLMs' proficiency in structuring tables and introduces a novel fine-tuning method, cognizant of data structures, to bolster their performance. We unveil \textsc{Struc-Bench}, a comprehensive benchmark featuring prominent LLMs (GPT-NeoX-20B, GPT-3.5, GPT-4, and Vicuna), which spans text tables, HTML, and LaTeX formats. Our proposed \textsc{FormatCoT} aids in crafting format-specific instructions from the intended outputs to populate this benchmark. Addressing the gap in task-centered evaluation, we propose two innovative metrics, P-Score (\textbf{P}rompting Score) and H-Score (\textbf{H}euristical Score), to more accurately gauge LLM performance. Our experiments show that applying our structure-aware fine-tuning to LLaMA-7B leads to substantial performance gains, outshining its LLM counterparts across most measures. In-depth error analysis and creating an ability map across six dimensions—coverage, formatting, reasoning, comprehension, pragmatics, and hallucination—highlight areas for future enhancements and suggest forthcoming research trajectories. Our code and models can be found at \url{https://github.com/gersteinlab/Struc-Bench}.

\section{Introduction}

Significant advancements have been made in various natural language processing tasks by Large Language Models (LLMs)~\cite{brown2020_gpt3, scao2022bloom, ouyang2022training, muennighoff2022crosslingual, openai2023gpt4,zhao2023survey}, especially in text generation tasks~\cite{qin2023chatgpt}. 
The ability to output structured data, one of the key aspects of generative capability, has also attracted great interest in previous studies~\cite{wu2021text,zhao2023robut,zhao2023large,zha2023tablegpt}.

\begin{figure*}[t]
    \centering
    \resizebox{0.9\linewidth}{!}{
    \includegraphics[width=1.2\textwidth]{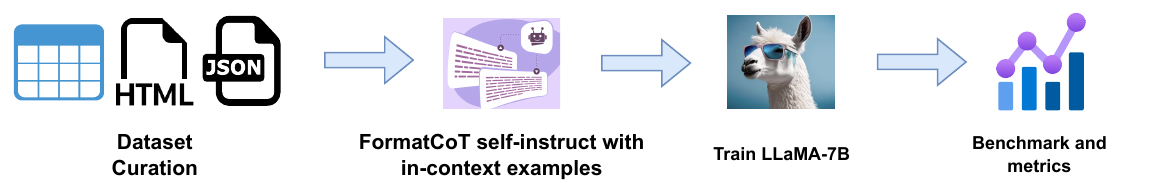}}
        \vspace{-0.3cm}
    \caption{Overview of our workflow: We commenced by creating datasets of raw text tables, HTML tables, and \LaTeX \ tables. Subsequently, LLaMA-7B was trained using the training data constructed by \textsc{FormatCoT}. Finally, our benchmarks validate the effectiveness of the current LLMs to generate such tables.}
    \label{fig:model}
    \vspace{-0.4cm}
\end{figure*}


Despite their advanced capabilities, LLMs have problems with generating complex structured tables, an indispensable skill for practical applications like coding copilot and automated report generation. This proficiency entails the organization of information from multifarious sources into coherent structures. Generating structured tables as outputs not only helps human understanding but also facilitates the automated data processing pipeline in autonomous language agents. Furthermore, generating structured tables can also serve as a critical preprocessing procedure for downstream tasks such as decision-making and knowledge extraction. However, the current landscape of LLM evaluation often neglects this aspect of table generation, which casts uncertainty on their full potential and utility in such scenarios. Our research seeks to thoroughly investigate these gaps.


\textit{First, there is a lack of systematic analysis and comprehensive benchmarks} of the ability of LLMs to output complex structured tabular data.
Previous efforts on evaluating LLMs \cite{qin2023chatgpt,ma2023large} on structured data primarily centered around simple Information Extraction (IE) tasks: recognizing named entities, extracting relations, and detecting events. Here the goal of IE tasks is to gather the extracted data in a highly structured form ~\cite{zhong2020frustratingly}. Much earlier work was considerably more task-centric as opposed to LLM-centric. The focus was predominantly on generating structured data from text (text-to-data) tasks with pre-trained  models~\cite{he2023revisiting,rossiello2022knowgl,whitehouse2023webie,pietruszka2022stable} like BART~\cite{lewis2019bart} and T5~\cite{raffel2020exploring}.

\textit{Second, there is a lack of evaluation metrics} of structured tabular data generation. Existing benchmarks often rely on rudimentary objective metrics such as word overlap to measure the accuracy of the content generated by the model~\cite{li2023sequence,wu2021text,pietruszka2022stable}. This may be insufficient for evaluating whether LLMs can generate structured output, as an ideal evaluation metric ought to also consider the format of generated content.

Third, there is a lack of methods to enhance the performance of current LLMs to better \textit{follow natural language inputs and generate tabular outputs with the correct format.}


Our contributions can be encapsulated as outlined in Figure \ref{fig:model}: (1) We introduce \textsc{Struc-Bench}, a benchmark specifically constructed for generating structured tabular data. (2) We evaluate popular LLMs on \textsc{Struc-Bench} using two proposed metrics, providing a comprehensive insight into the prevailing limitations and common error types.
(3) We propose \textsc{FormatCoT} to generate instruction tuning data, wherein we utilize GPT-3.5 to generate format instructions and then fine-tune LLaMA-7B model to follow these formats. The resulting impressive performance demonstrates that with \textsc{FormatCoT} small models can indeed surpass the performance of a larger model in this particular task.

\vspace{-0.1cm}

\section{Problem Analysis and Benchmark}

\vspace{-0.1cm}

\subsection{Problem Definition and Motivation}

\vspace{-0.1cm}

LLMs are tasked with generating complex structured tables, a process that involves understanding both the content and the specific format requirements, such as LaTeX syntax. This task extends beyond simple text generation as it demands precision not just in content creation but also in adhering to a detailed and precise structural format. Specially, we aim to convert unstructured textual data into structured tabular data, by extracting necessary contents from text and following the required structure or format.

\subsection{Problem Analysis}

To assess LLMs' capability to convert textual descriptions to structured tables, we utilized the RotoWire dataset \cite{wiseman2017challenges}, originally a table-to-text dataset, in reverse as a text-to-table task. After ensuring that the descriptions contained adequate information for table generation through a review of 20 samples, we found significant limitations in the performance of GPT-3.5 and GPT-4, especially when dealing with complex structures as detailed in Appendix \ref{exampleA}.

\begin{figure*}[t]
    \centering
  \includegraphics[scale=0.5]{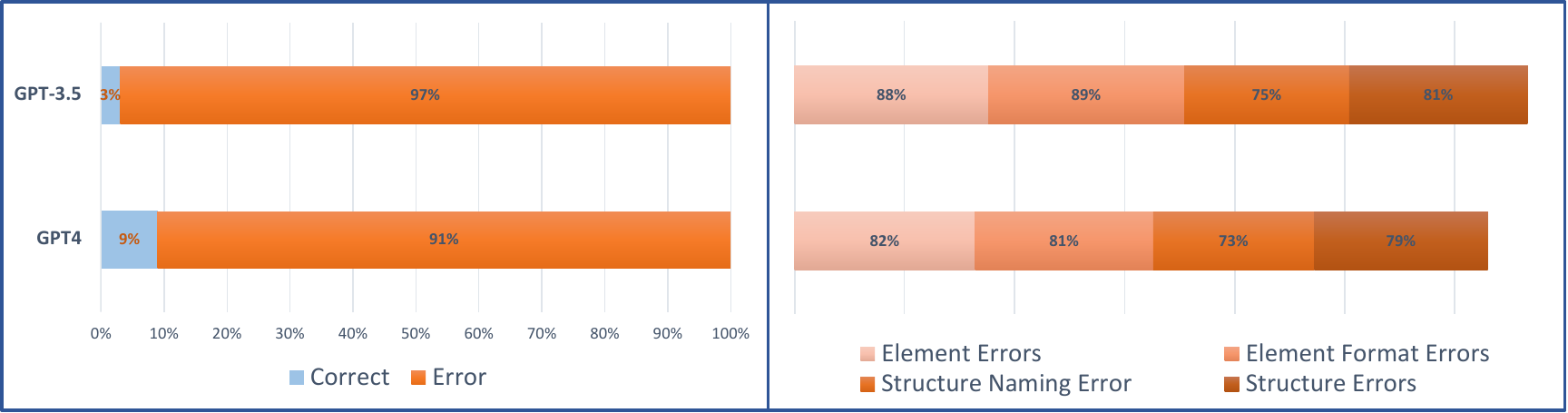}
    \caption{Error analysis by human annotation. Some error types are explained in Appendix \ref{exampleA}.}
\vspace{-0.3cm}
    \label{fig:error}
\end{figure*}

When put to the test of generating data in precise formats, such as tables, both GPT-3.5 and GPT-4, despite their advanced capabilities, frequently erred, as evidenced by a systematic MTurk human annotation study (refer to Appendix \ref{mturk}). The types of errors, categorized into `Element Errors', `Element Format Errors', `Structure Error', and `Structure Naming Errors', are quantified in Figure \ref{fig:error}.

A mere 3\% of GPT-3.5's outputs were fully accurate, with GPT-4 only slightly better at 9\%. These results suggest design limitations within the GPT architecture, which, although effective at mimicking language patterns, falter in tasks requiring sustained structural coherence over longer sequences.

\vspace{-0.1cm}

\begin{figure*}[b]
    \centering
           \vspace{-0.3cm}
\includegraphics[width=0.9\textwidth]{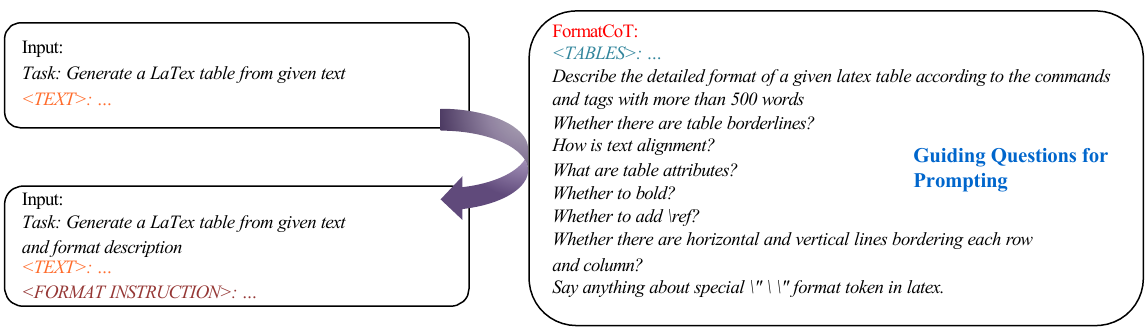}
        \vspace{-0.2cm}
    \caption{The upper-left corner box represents the original input, which notably lacks a description of the format. To explicitly instruct the model on format understanding, we employ the \textsc{FormatCoT} located on the right, which produces the <FORMAT INSTRUCTION>. The lower-left box illustrates what the input for LLaMA fine-tuning looks like after passing through \textsc{FormatCoT}. <TEXT> provides a descriptive text for the expected table output (original input), <TABLE> serves as a reference table (output), and the <FORMAT INSTRUCTION> is a format guideline generated through \textsc{FormatCoT} (added into input). Detailed prompts are displayed in Appendix \ref{FormatCoTPrompt}.}
    \label{fig:FormatCoT}
\end{figure*}

\subsection{Benchmark Construction}

\begin{figure}[t]
    \centering
    \includegraphics[width=0.47\textwidth]{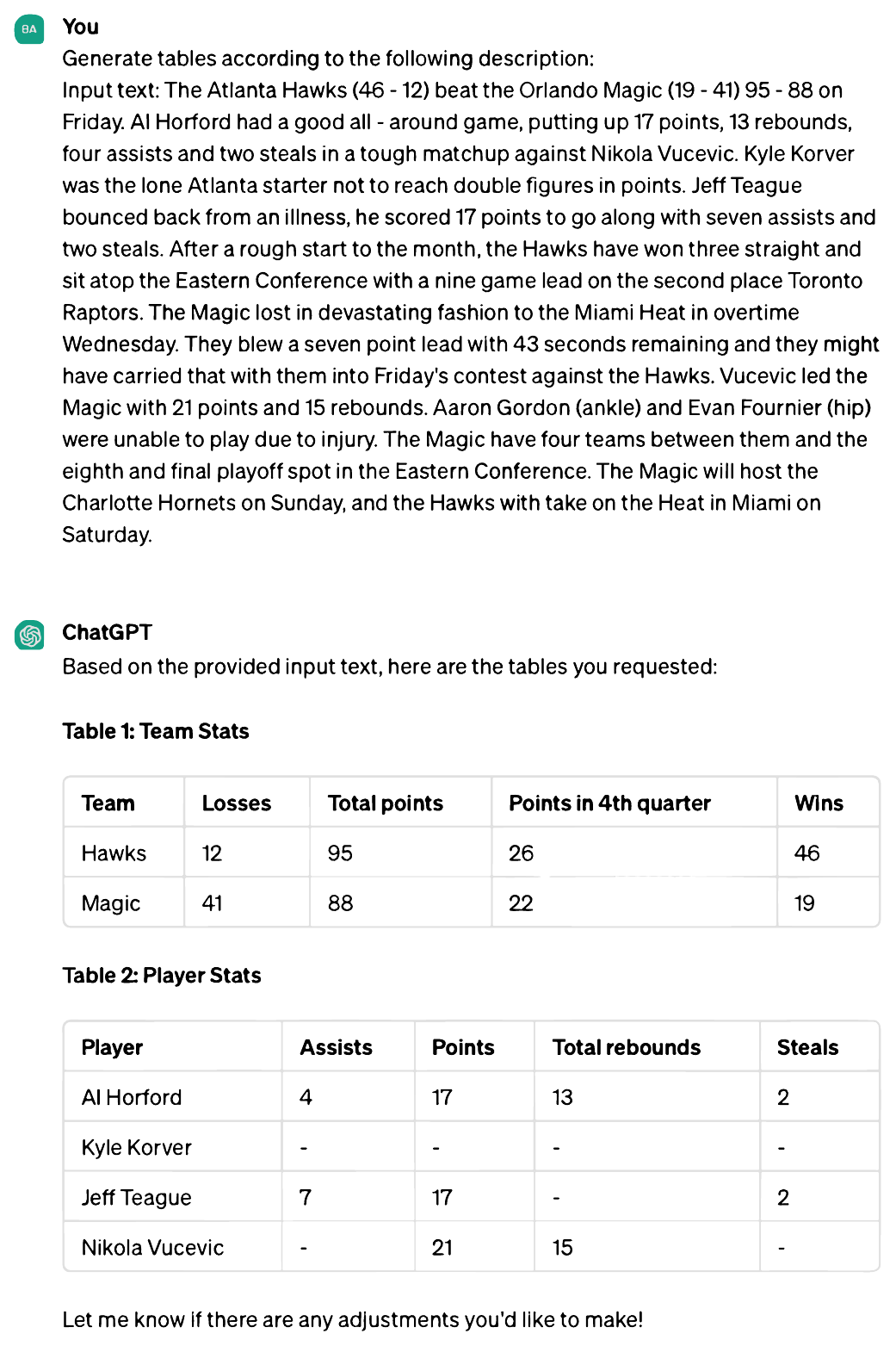}
\caption{An exemplification of our benchmark. The input is made up of the instruction and the input text, whereas the output aims to present the target table. Notably, there are some inaccuracies in the predicted output; for instance, `Points in 4th quarter' under `Hawks' should be vacant, and correspondingly, `Points in 4th quarter' for `Magic' should be 21.}
    \label{fig:chatgpt}
    \vspace{-.4cm}
\end{figure}

We begin by selectively sourcing tables larger than 3x3 from the RotoWire \citep{wiseman2017challenges} dataset to present a baseline of complexity. 
Then, to broaden our dataset diversity across various domains, from The Stack \citep{Kocetkov2022TheStack}, which includes GitHub code in 358 programming languages from the BigCode project, we first select LaTeX and HTML formats. Further refining our dataset, we extract elements relevant to table representations to ensure focused complexity and relevance to our structured data generation task. An example of our benchmark is shown in Figure \ref{fig:chatgpt}.

Table \ref{tab:dataset} gives statistics for the Rotowire dataset and our constructed datasets.
Then we evaluate 4 popular LLMs, including GPT-NeoX-20B~\cite{black2022gpt}, GPT-3.5, GPT-4, and Vicuna-13B~\cite{chiang2023Vicuna}.
For LaTeX and HTML data without paired text, we harness GPT-3.5 to construct synthetic descriptions to be utilized as input. 
To guarantee the quality of our benchmark, we sample 50 tables for each format to ensure the correctness of the descriptions. Initially, we achieved a satisfaction rate of 76\%. However, upon incorporating a manual interpretation template (e.g. tab names for HTML) tailored to each format (Appendix \ref{PromptDescription}), our satisfaction rate improved significantly, reaching 96\%.
For example, HTML tables possess their unique tags and structure, conforming faithfully to the syntax rules of HTML language.

\begin{table}[h!]
    \centering
    \resizebox{\linewidth}{!}{
    \begin{tabular}{lccccc}
        \toprule
        \textbf{Dataset} & \textbf{\# Train} & \textbf{\# Test} & \textbf{Format} & \textbf{ Rows \& Columns}\\
        \midrule 
        \textsc{Struc-Bench} Table & 3.4k &  728 & Raw tex & 7.26 \& 8.75  \\
        \textsc{Struc-Bench} \LaTeX & 5.3k &  500 & \LaTeX &  2.75 \& 4.47  \\
        \textsc{Struc-Bench} HTML & 5.4k &  499 & HTML & 5.50 \& 3.54 \\
        \bottomrule
    \end{tabular}
}
\caption{\textsc{Struc-Bench} data statistics. The number of Rows \& Columns has been averaged.}
    \label{tab:dataset}
    \vspace{-0.4cm}
\end{table}

\begin{table*}[b!]
\centering
\resizebox{\linewidth}{!}{
\begin{tabular}{lccccccccc}
\toprule\toprule
Model & SacreBLEU & ROUGE-L & BERTScore & BARTScore & BLEURT&  Content P-Score & Format P-Score & Content H-Score & Format H-Score \\
\midrule
\multicolumn{10}{c}{\emph{Tables from Raw Text}}\\
GPT-NeoX-20B &35.24&55.78&68.91&-2.34&33.51&3.86&6.10&0.50&-1.32 \\

GPT-3.5 & 56.92 & 70.97 & 91.35 & -1.68 & 36.85 & 6.19 & 8.16 & 0.52 & -1.27 \\
GPT-4  &68.13&75.44 &94.89 &-0.99 & 55.24& 6.88& 8.30& 0.85& 0.53\\
Vicuna-13B &40.12& 50.77&75.21 & -2.05& 40.02&4.07 &6.33 & 0.55 & -1.38\\
 Ours-7B & \textbf{90.6} & \textbf{88.98} & \textbf{98.54} & \textbf{-0.69} & \textbf{66.07} & \textbf{7.69} & \textbf{8.60} & \textbf{1.65} & \textbf{3.61} \\
 
\quad $w.o. finetune$ & 9.9 & 36.56 & 81.63 & -2.50 & 70.24 & 4.58 & 6.00 & 0.51 & -1.01 \\

\midrule
\multicolumn{10}{c}{\emph{LaTeX}}\\

GPT-NeoX-20B & 45.92 & 65.10  & 76.09& -2.05 & 40.87 &7.23 & 7.02&0.56 &0.72 \\

 GPT-3.5 & 56.94 & 75.99  & 86.25& -1.30 & 42.89 &8.22 & 8.41&0.99 &1.27\\
 GPT-4  & 78.15 & 85.34  & 88.07& -1.09 & \textbf{67.11}&8.78&8.81&1.10&1.35 \\
Vicuna-13B &50.80 & 69.48  & 80.44& -1.07 & 36.74 &7.70 & 8.10&0.78 &1.06 \\
 Ours-7B & \textbf{89.13} & \textbf{88.99} & \textbf{98.55} & \textbf{-0.69}& 66.07& \textbf{8.94}& \textbf{9.05}& \textbf{1.14}& \textbf{1.52} \\
 
\quad $w.o. finetune$ &47.24 & 70.89  & 73.27& -2.13 & 38.13 &7.10 & 6.98&0.51 &0.69 \\

\midrule
\multicolumn{10}{c}{\emph{HTML}}\\
GPT-NeoX-20B &60.36&72.13&86.88&-1.59&30.06&8.42&8.94&0.81&0.92 \\

GPT-3.5 & 73.80 & 85.19 & 96.76 & -1.46 & 34.81 & 9.11 & 9.35 & 1.10 & 2.15 \\
GPT-4  &\textbf{79.25}& 85.95  &\textbf{97.22} &-1.31 &41.59 &9.17 &9.62 &1.15 &2.29 \\
Vicuna-13B &58.75&70.37&88.65&-1.58&31.11&8.55&8.88&0.79&0.93 \\
Ours-7B  & 77.50 & \textbf{86.08} & 96.25 & \textbf{-1.30} & \textbf{42.89} & \textbf{9.20} & \textbf{9.70} & \textbf{1.18} & \textbf{2.49} \\
 
\quad $w.o. finetune$ & 65.30 &  78.24 &88.12&-1.57&32.78&8.22&8.81&0.92&0.96 \\

%
 

\bottomrule

\bottomrule
\end{tabular}
}

\caption{Automated evaluation results on the \textsc{} test set, involving five types of previous metrics and four proposed ones.
$w.o. finetune$ means that we also compared the performance of our model without finetuning as an ablation study. 
`Ours-7B' is finetuned LLaMA.}
\vspace{-0.5cm}
\label{tab:automated_evaluation_result}
\end{table*}

\vspace{-0.2cm}

\section{Methodology}

\vspace{-0.1cm}

\subsection{Data Generation}
As shown in Figure \ref{fig:FormatCoT}, we propose \textsc{FormatCoT} with GPT-3.5, a self-instruct method to generate \{data, instruction\} pairs for fine-tuning purposes.
Specifically, our prompt of \textsc{FormatCoT} involves guiding models to accurately describe and interpret the format elements presented in the output table, inspired by \citet{wang2023element} in the summarization task. To verify the effectiveness of our proposed \textsc{FormatCoT}, we conduct an ablation study in Appendix \ref{ablation}.

\vspace{-0.2cm}
\subsection{Instruction Tuning}
We introduce an instruction tuning approach designed specifically to enhance LLMs' abilities in generating structured text \cite{touvron2023llama,gorilla}. Specifically, we combine GPT-3.5-generated format descriptions of output tables and the original text input as the new input of LLaMA fine-tuning. In other words, we start with GPT-3.5 processing table data and synthesizing comprehensive format instructions. The LLaMA model is then fine-tuned on these enriched instructions we generate.
This approach simulates a user-agent interaction where GPT-3.5 effectively fetches and consolidates table information, conversationally instructing LLaMA for the final text generation, outlined in Figure \ref{fig:FormatCoT}.


\vspace{-0.2cm}

\subsection{Evaluation Metrics}

Assessing the accuracy of generated tables against ground truth is complex due to the variability in formatting, like HTML. An ideal evaluation metric needs to discern substantial data discrepancies while disregarding trivial formatting variations.

We propose to break down the similarity of two tables into two coarse components: \textit{content} and \textit{format}. 
In scoring \text{content} similarity, we attempt to parse \textit{content} out the data within the table cells, and compute the similarity. This similarity is computed between the generated and ground-truth table cells by commonly used similarity metrics.
In scoring \text{format} similarity, we place higher emphasis on components such as the number of columns and rows, cell alignment, and the table caption. 
We find that these two scores allow us to perform a more involved analysis of where predicted and ground-truth tables differ.
The implementation of these two scores can be found in Appendix \ref{scoring}.

\vspace{-.1cm}
\subsubsection{P-Score}

We take two approaches to score each metric. 
First, we perform model-based evaluation, querying GPT-3.5 with both tables and having it score the similarity of content and format separately.
Following \citet{wang2023large}, we prompt the model to perform Chain-of-Thought \cite{wei-2023-chain-of-thought} reasoning before outputting its scores, and we query the model with the predicted and ground-truth tables in both orders and average the scores.
We report these as the \textit{P-Score} (Prompting Score).

\vspace{-0.1cm}

\subsubsection{H-Score}

In addition, we also implement hand-crafted scoring functions to score the similarity of the tables. 
Since the tables can be presented in different formats, we implement several heuristics to normalize the tables and to compute their similarity. We use an average of Levenshtein distance and the Ratcliff/Obershelp similarity metric to compute the similarities between strings or data structures. These heuristically normalized metrics are reported as the \textit{H-Score} (Heuristical Score).
The analysis can be found in Appendix \ref{validity}.

\vspace{-0.1cm}

\section{Experiments}

\vspace{-0.1cm}
\subsection{Basic Settings}
\label{settings}
For metrics, we use SacreBLEU, ROUGE-L, BERTScore, BARTScore, and BLEURT metrics as they are all classical metrics to evaluate text similarity, as well as two proposed metrics: P-Score and H-score. 
qWe evaluate the following models: GPT-NeoX-20B, GPT-3.5, GPT-4, Vicuna-13B, LLaMA-7B, and our finetuning LLaMa-7B. 
GPT-NeoX-20B, GPT-3.5 and GPT-4 represent the state-of-art performance of current LLMs and Vicuna-13B is another version finetuned on LLaMA, which can reach 90\% of the capacity of GPT-3.5. We think these models are strong enough to be persuasive. 
For the first 4 models, we simply call their APIs from OpenAI or HuggingFace to generate results without further finetuning. 
In our dataset, each item consists of three parts: instruction, input, and output. When generating results, we put each item's instruction and input together as the final input to models.
During inference, the user provides the prompt in natural language, this can be for a simple task (e.g., ``please generate a table given by the following information and format'').
During the inference process, we provide the model with a natural language prompt to describe the format and content of our task, as well as the expected response.
\vspace{-0.1cm}

\subsection{Human Evaluation}

Table \ref{tab:human_evaluation_results} displays human evaluation results on two proposed metrics with instance-level Pearson correlation, reflecting a purposeful design that caters to the specific demands of structured output assessment. We engaged five undergraduate students to annotate 200 examples focusing on content and format quality. Equipped with the input description (with reference appended) and generated outputs, they scored each aspect on a 10-point scale. Both the P-score and H-score showcase a significant correlation with human judgment, indicating their relative robustness and effectiveness in this evaluation space. This level of correlation, which surpasses that of many prior meta-evaluation efforts \cite{fabbri2020summeval,tang2021investigating}, reinforces the value of our metrics and addresses concerns about their ability to reliably reflect human evaluation.
Additionally, we evaluated well-known metrics including ROUGE-L, BERTScore, BARTScore, and BLEURT. Limited space precluded a full discussion, yet our Content P-score showed the best instance-level correlation.

\label{human evaluation}
\begin{table}[h!]
    \centering
    \resizebox{0.98\linewidth}{!}{
    \begin{tabular}{lcc}
        \toprule
        \textbf{Metrics} & \textbf{Content Correlation} & \textbf{Format Correlation}\\
        \midrule 
        Content P-score & 0.5301 & - \\
        Format P-score & - & 0.3812 \\
        Content H-score & 0.1059 & - \\
        Format H-score & - & 0.3021 \\
        \bottomrule
    \end{tabular}
}
\vspace{-.1cm}
\caption{Human evaluation results.}
\vspace{-0.4cm}
    \label{tab:human_evaluation_results}
\end{table}

\subsection{Results}
\label{results}

\begin{figure}[h!]
    \centering    \vspace{-0.05cm}
    \includegraphics[width=0.46\textwidth]{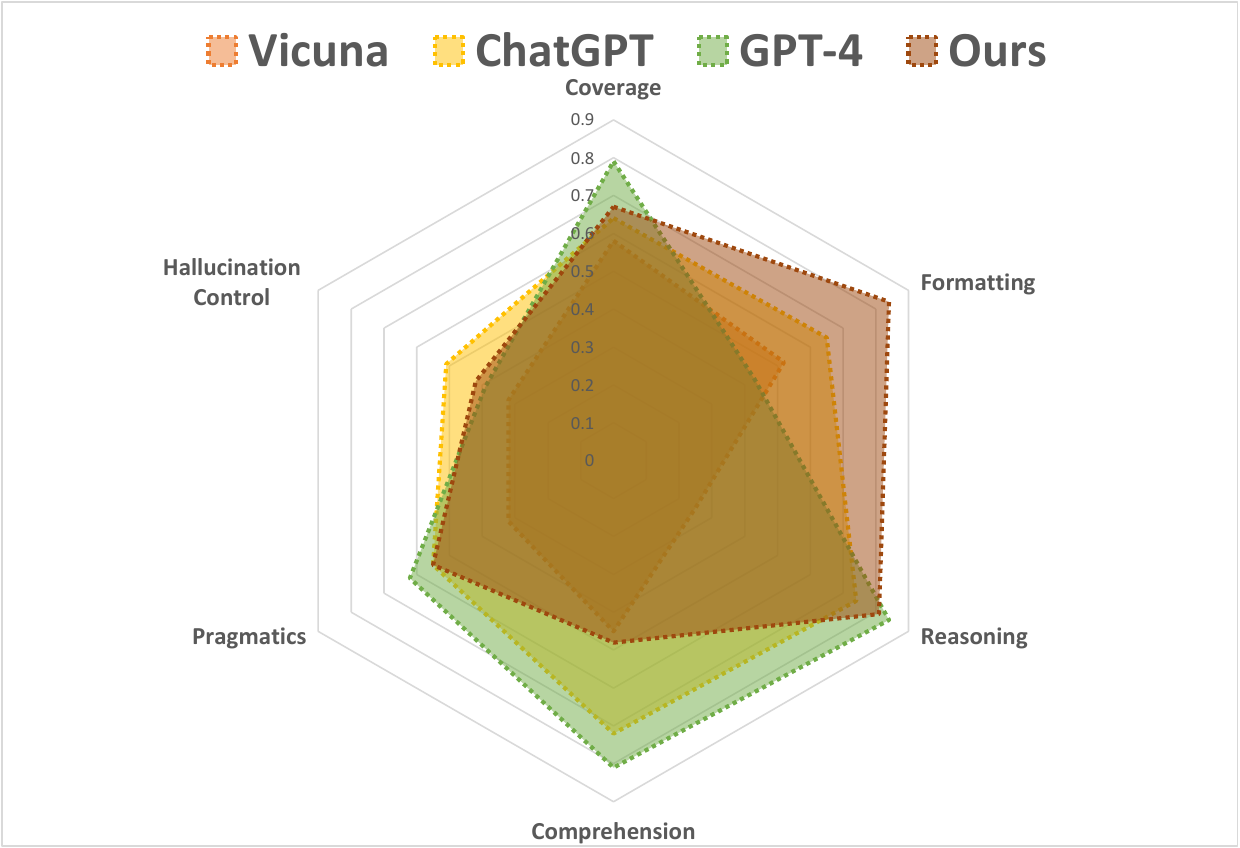}
    \caption{Visualization of LLMs' capability.}
    \label{fig:visual}
\vspace{-0.3cm}
\end{figure}

Table \ref{tab:automated_evaluation_result} provides a comparative analysis of different LLMs based on several metrics. For `Tables from Raw Text', the Ours-7B outperforms the other models in every metric. Interestingly, without fine-tuning, the performance drops significantly, particularly in SacreBLEU, ROUGE-L, and BERTScore. 
The results for `LaTeX' reveal a similar trend and in the `HTML' category, GPT-4 scores the highest in SacreBLEU and BERTScore. However, these differences are slight and our 7B model comes out on top for the rest of the metrics.
The results demonstrate that our approach exhibits superior performance, highlighting the efficacy of fine-tuning smaller models in surpassing much larger models.
Moreover, we delve into an analysis based on our Mturk annotation, attributing observed shortcomings to several error types. And we present an ability map in Figure \ref{fig:visual} and Appendix \ref{map}. 



\section{Conclusion}

\vspace{-.1cm}
In summary, our study provides a thorough analysis of LLMs' challenges in structured table generation, introduces novel evaluation metrics, and assembles a specific benchmark covering a range of data types. We pinpoint key issues including content fidelity, format adherence, numerical reasoning, and management of extensive tables.

\clearpage
\section{Limitations}

Although we present a comprehensive analysis, the exploration of LLMs in structured text generation presented in this paper has several limitations:

\paragraph{Investigating Optimal Format for Tabular Representation} In this study, we did not investigate which table formats are most effective. Different presentations of the same information can be reasonable, and \textbf{table normalization} strategies, such as determining the best way to tabulate given facts or how to interconnect multiple tables, remain unexplored. Future research could engage in the study of table normalization to ascertain optimal strategies for tabular data structuring and representation.

\paragraph{Domain-Specific Benchmark Development} While we've made strides in constructing benchmarks for structured text generation, it may be beneficial to develop benchmarks that cater to specific domains. Different fields might have unique structural requirements and understanding these nuances can significantly improve the models' applicability across diverse contexts.

\paragraph{Expand the Range of Datasets} There are endless data types and sources that can be explored. Incorporating a broader variety of datasets could expose the models to an even wider range of structural formats, ultimately enhancing their overall performance.

\paragraph{Enhancing Numerical Reasoning Capabilities} Our study identified inadequate numerical reasoning as one of the challenges faced by LLMs. Investigating techniques to bolster numerical reasoning in these models could lead to significant improvements in their performance.

\paragraph{Developing Advanced Methods} While our structure-aware instruction tuning method showed promising results, more sophisticated techniques could be developed. For instance, future work could explore ways of incorporating more explicit structural information into the model or developing methods that allow the model to learn structural patterns more effectively.

\paragraph{Exploring Multimodal LLMs} As LLMs continue to evolve, there are opportunities to explore multimodal models that can process and generate both text and other forms of data, such as sound or images~\cite{kamigaito2023table}, in a structured manner.

\bibliography{anthology,custom}
\bibliographystyle{acl_natbib}

\clearpage
\appendix

\section{Analysis with Examples}
\label{exampleA}

\subsection{Example Table A}
\label{exampleA1}

\begin{figure*}[!hb]
    \centering
    \includegraphics[width=0.64\textwidth]{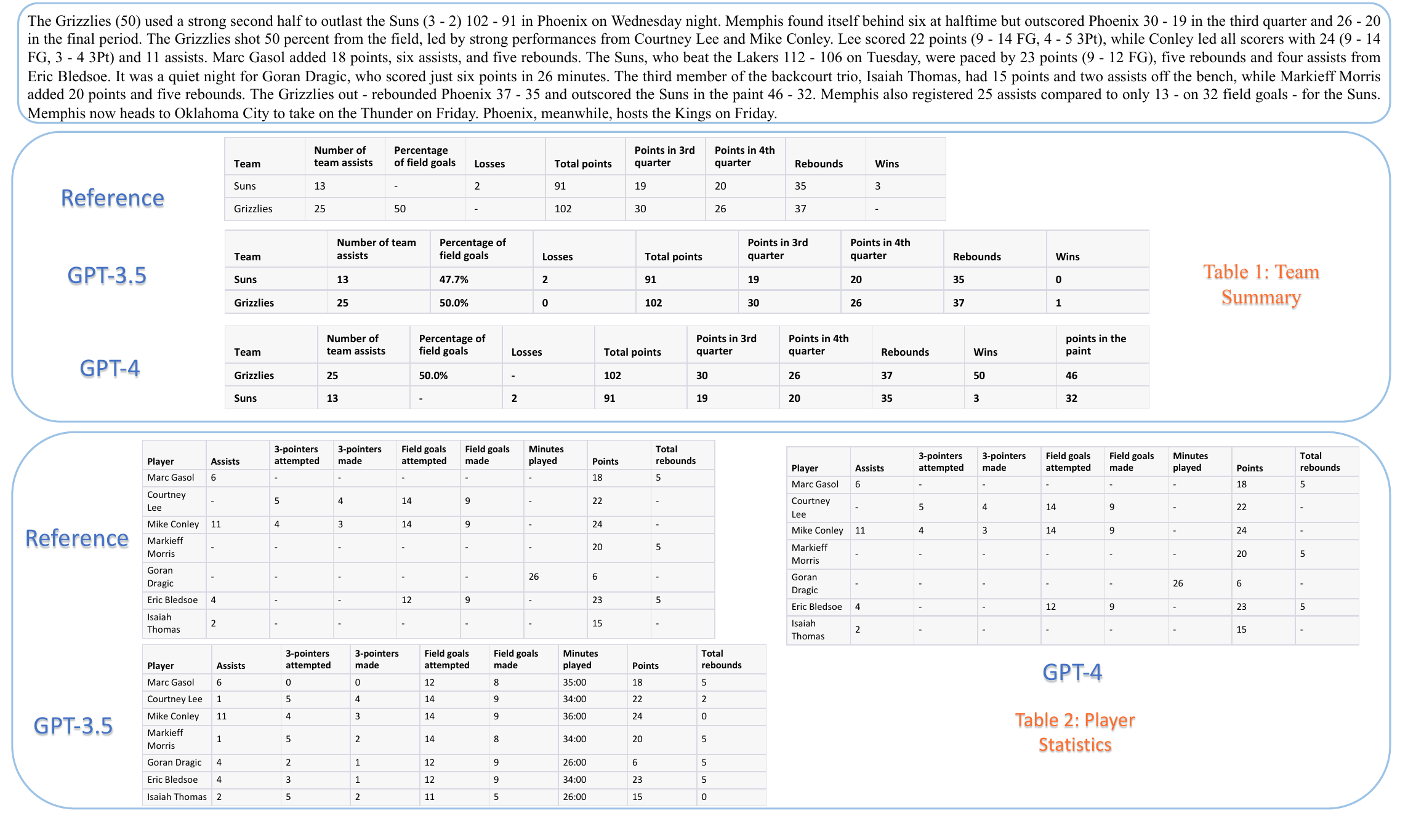}
    \caption{Examples of generating tables with GPT-3.5 and GPT-4 based on the input text. The generated results contain a large number of errors, including format errors and content errors.}
    \label{fig:example}
\end{figure*}

The main difference between the reference tables and the tables generated by GPT-3.5 and GPT4, shown in figure \ref{fig:example}, is in the completeness and precision of the data provided.

In the reference tables, all relevant data is fully represented:
For the teams (Table 1), each team has a precise number or percentage for every statistic. Similarly, for the players (Table 2), each player has a definite number for every statistic, including minutes played in the format ``mm:ss''.

In contrast, the generated tables show data that is incomplete and imprecise. For GPT-3.5 generated one, the team statistics table has some statistics missing, as represented by empty cells, and some are not presented as percentages. The player statistics table also has missing data similarly, and it lacks the "minutes played" statistics entirely. For instance, in the `team' table, the "Percentage of field goals" column for the Suns is missing. Similarly, in the `player' table, many key statistics such as "3-pointers attempted", "3-pointers made", "Field goals attempted", "Field goals made", and "Minutes played" are missing for various players.
Regarding the format, we observe a lot of format errors. For example, the `Percentage of field goals' column for Grizzlies is represented as "50" instead of "50.0\%". Moreover, the `Wins' column for the Suns is represented as "3" instead of "0". This misrepresentation can lead to significant misunderstanding of the data.
The `Player' table also has format errors. For instance, the `Minutes played' column is missing the time format (i.e., ``00:00'').
On the other hand, the reference tables adhere to a standard format. Percentage data is represented with a `\%' sign, time data uses the  `00:00' format, and numeric data correctly represents each statistic. 

\begin{figure*}[!ht]
    \centering
    \includegraphics[width=0.6\textwidth]{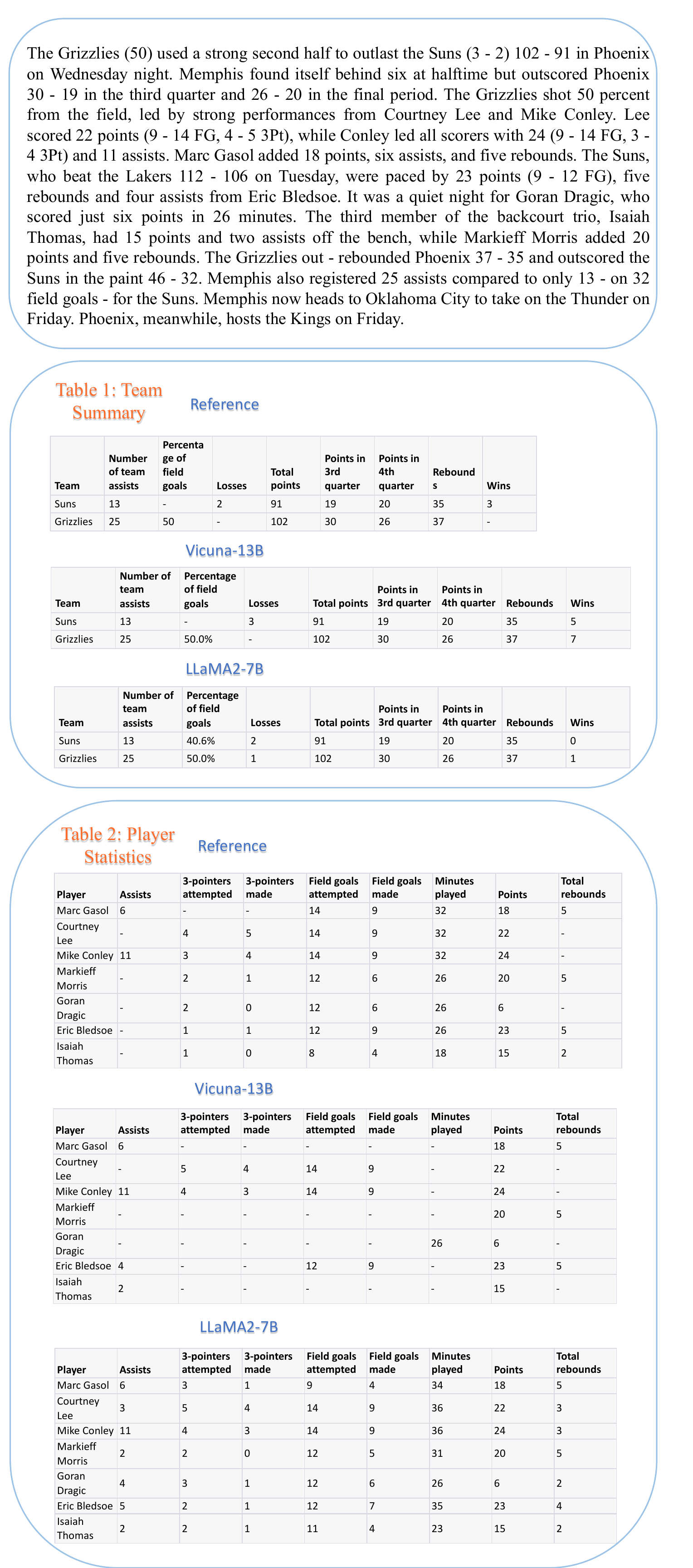}
    \caption{Examples of generating tables with Vicuna-13B and LLaMA2-7B based on the input text, the generated results contain a large number of errors, including format errors and content errors.}
    \label{fig:example2}
\end{figure*}

For Vicuna-13B results shown in figure \ref{fig:example2}, although it has the correct format for both tables, there are still many element errors. For instance, the `team' table has wrong statistics such as ``Losses'' and ``Win'' for the Suns. Besides, in the `player' table, many cells shouldn't have data. However, they have, which is a mistake. Some cells like Isaiah Thomas's and Eric Bledsoe's `Assists' should be 2 and 4, but they are not in the Vicuna-13B `player' table. Similarly, LLaMA2-7B results, have the same element errors in the `team' table and worse errors in the `player' table. It fills all cells, many of which should be none. As for some cells that should have data, their data are wrongly filled in like Eric Bledsoe's `Assists' and `Field goals made'.

\begin{figure*}[!hb]
    \centering
    \includegraphics[width=0.64\textwidth]{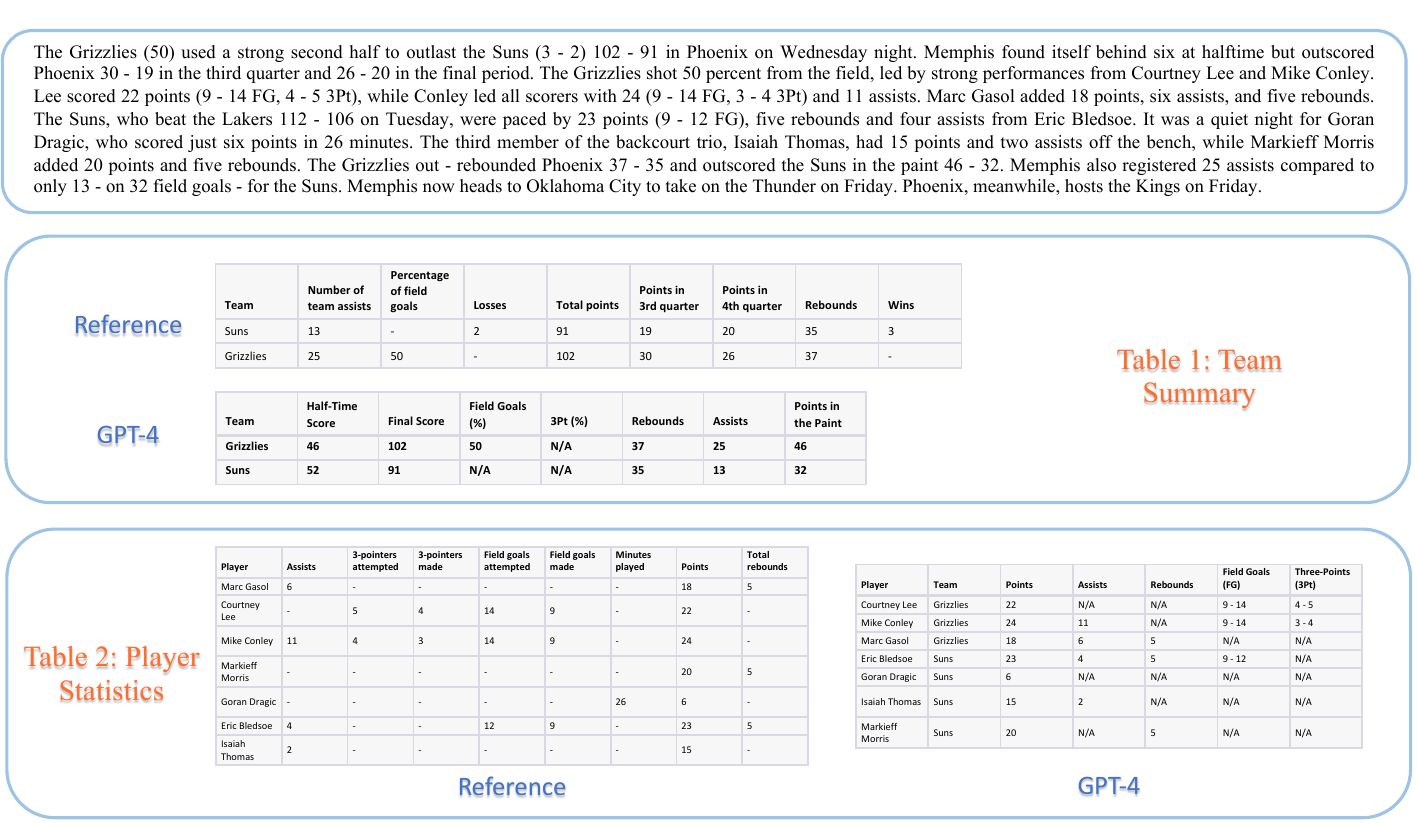}
    \caption{Using GPT-4 to generate a table based on the input text without \textsc{FormatCoT}, the generated results contain a large number of errors, including format errors and content errors.}
    \label{fig:errors}
\end{figure*}

\clearpage

\clearpage

\subsection{Error Type}
\label{errortypeA}

\paragraph{Structure Errors:} These errors pertain to the structural integrity of the generated tables. Specifically, they include instances where there are excess or missing rows or columns in comparison to the correct table structure. For instance, in figure \ref{fig:errors} GPT4 generated result has missing columns like ``Win'' and ``Losse'' in the `team' table.

\paragraph{Structure Naming Errors:} This category captures errors related to the naming conventions used for rows or columns. Any discrepancies in a row or column names between the generated and correct table are flagged as structure naming errors. For instance, in figure \ref{fig:errors}, the GPT-4 generated result has wrong column names like ``Half-Time Score'' in the `team' table.

\paragraph{Element Errors:} These are inaccuracies observed at the element level within the generated table. Element errors encompass incorrect numbers, values, or inappropriately empty cells, reflecting discrepancies in individual table entries relative to the correct table. In figure \ref{fig:example} and figure \ref{fig:example2}, most errors are element errors.

\clearpage
\subsection{Metric Effectiveness}
\label{validity}

Table \ref{tab:metrics} crystallizes different results of examples in Appendix \ref{exampleA1} based on our H-score metric. Taking GPT-4 as an example, there exist slight drops from 2.0 to 1.86 in content H-score, and very accurately, this trend is followed by the results of GPT-4 revealing slight errors in its content. For the format H-score, GPT-4 also cannot do very well for its numeric performance, which is not close enough to the full score, matched by the poor performance of GPT-4 in this area. Other models' H-scores follow this trend as well. Therefore, the H-score can respond to differences in both content and format accurately.

\begin{table}[h!]
    \centering
    \resizebox{\linewidth}{!}{
    \begin{tabular}{lccccc}
        \toprule
        \textbf{Model} & \textbf{Content H-score} & \textbf{Format H-score}\\
        \midrule 
        GPT-3.5 & 1.39 & 4.0 \\
        GPT-4 & 1.86 & 3.44 \\
        Vicuna-13B & 1.51 & 4.0 \\
        LLaMA2-7B & 1.37 & 4.0 \\
        Ours-7B & 2.0 & 4.0 \\
        Reference & 2.0 & 4.0 \\
        \bottomrule
    \end{tabular}
}
\caption{H-scores for different results of Ours-7B, GPT-3.5, GPT-4, Vicuna-13B and LLaMA2-7B.}
\vspace{-0.4cm}
    \label{tab:metrics}
\end{table}

\clearpage






\clearpage

\section{MTurk}
\label{mturk}
To facilitate a comprehensive analysis of the LLM output, we designed a task on Amazon Mechanical Turk (MTurk) to gather detailed annotations about various error types encountered in the generated structured tables. The task was structured as follows:

From the RotoWire dataset, we randomly chose 100 instances of LLM-generated output, ensuring a representative mix of quality based on preliminary assessment.

Each Human Intelligence Task (HIT) presented the annotators with a side-by-side view of the LLM output and the expected structured table format. Annotators were instructed to identify and categorize errors according to predefined types: `Element Errors', `Element Format Errors', `Structure Error', and `Structure Naming Errors'.

We provided extensive guidelines, exemplified with step-by-step instructions, to clarify typical instances of each error type. These guidelines were reviewed and iteratively improved through a pilot study conducted with a small set of annotators.

About the qualifications of Amazon Mechanical Turk (MTurk) workers, we use the following qualifications to recruit in total of 10 MTurk workers with good track records: HIT approval rate greater than or equal to 98\%, number of HITs approved greater than or equal to 500, and located in one of the following English native-speaking countries: Australia, Canada, New Zealand, United Kingdom, United States.
Each annotator is limited to annotating 10 examples, including both the output of GPT-3.5 and GPT-4.

Annotators workers were compensated \$7, calibrated to equal a \$42/hour pay rate. We first annotated examples in-house to determine the required annotation speed. A summary block usually takes around 10 minutes.

To demonstrate our annotation template and facilitate future research, we show the interface for annotations.

\begin{figure}[h]
    \centering
    \includegraphics[width=0.4\textwidth]{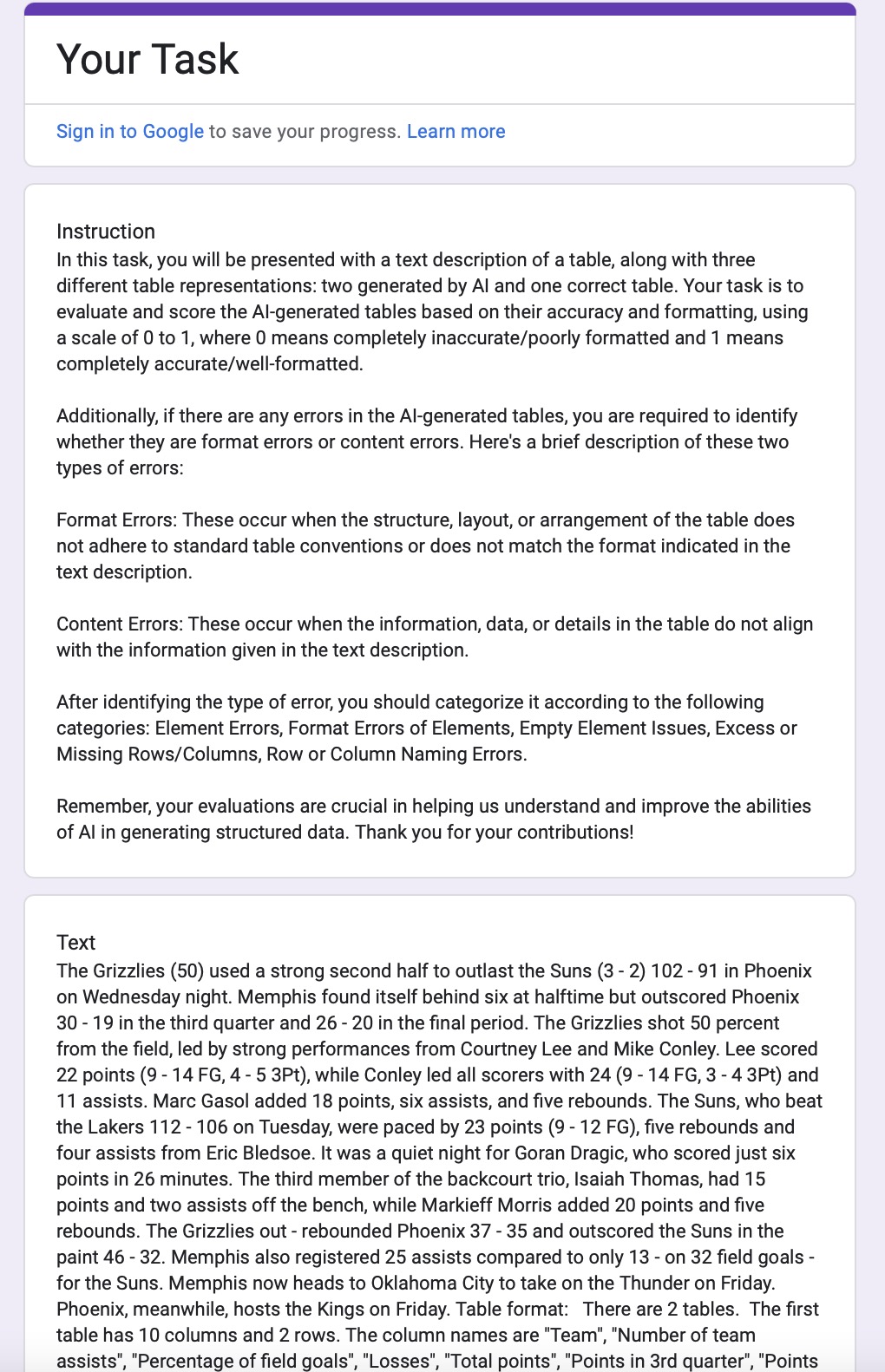}
    \caption{Our interface of Mturk.}   
    \label{fig:mtruk1}
\end{figure}

\begin{figure}[h]
    \centering
    \includegraphics[width=0.4\textwidth]{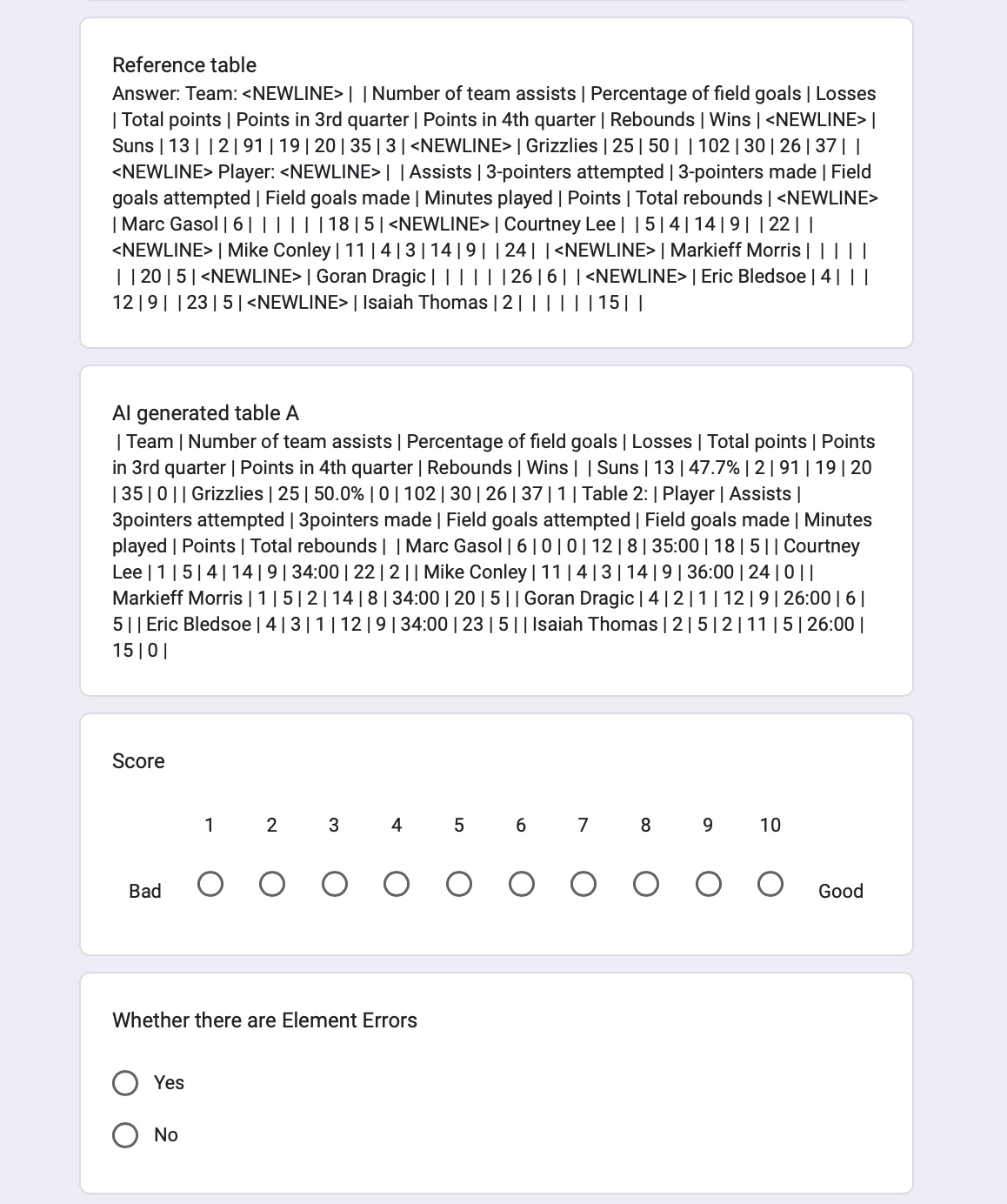}   
    \caption{Our Interface of Mturk.}   
    \label{fig:mtruk2}
\end{figure}

\clearpage

\section{Scoring}
\label{scoring}
\subsection{P-Score}
\label{P-Score}
Our approach involves prompting the model to engage in Chain-of-Thought reasoning before issuing its scores. Firstly, we instruct GPT on how to evaluate both "content similarity" and "structure similarity". Following this, the model is guided on the correct procedure to output its answer. To calculate the scores, the model is queried with both the predicted table and the ground truth table in varying sequences, after which the scores are averaged. We'll illustrate this process using the P-Scores prompt for raw text tables as an illustrative example:

\textit{
“Based on the above, we wanted to determine if the above tables are similar. Ideally, they should have identical content and structure. Score the "content similarity" and "structural similarity" between 0 and 10.}

\textit{
- Content similarity: 10 if the contents of the table cells are identical, 0 if they are entirely different. If about 50\% of the cells have the same data, the score should be 5.}

\textit{
- Structural similarity: 10 if the tables have the same structure (e.g. same column and rows with identical ordering, same alignment, etc.) although text formatting differences can be ignored (e.g. colors, font).}

\textit{Output a JSON object such as the following:}

\textit{"""json}

\textit{\{\{}

\textit{  "content_similarity": ...}
  
\textit{  "structural_similarity": ...}

\textit{\}\}}

\textit{"""}

\textit{Think carefully, and then output the scores.”
}

\begin{figure*}[t]
    \centering
    \includegraphics[width=1\textwidth]{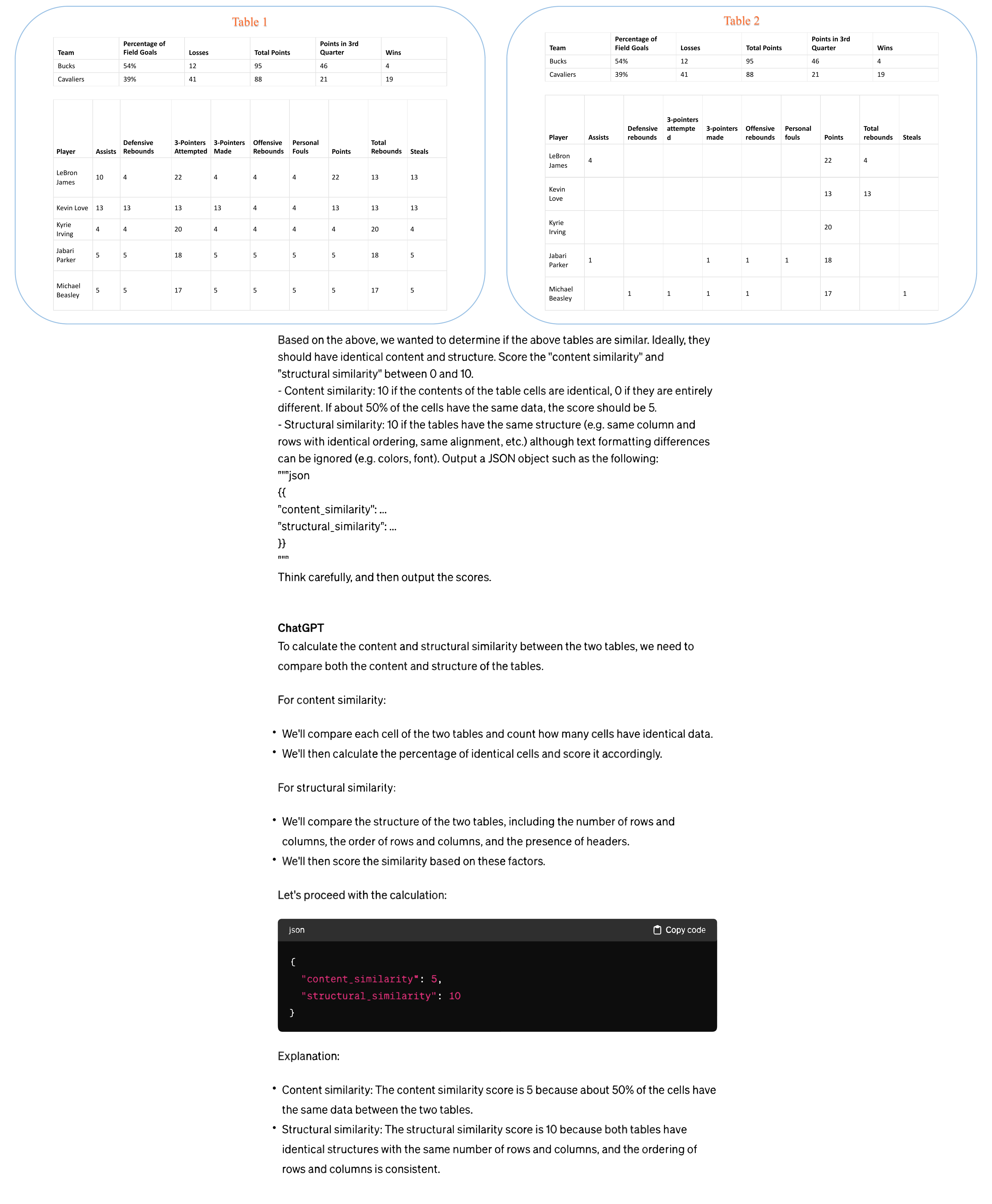}
    \caption{An example of P-score calculation. The top two figures displays two tables, accompanied by a prompt below to calculate the content and structural similarity between them. The goal is to determine if the tables have identical content and structure. The content similarity will be scored based on the percentage of cells with identical data, ranging from 0 (no similarity) to 10 (complete similarity). The structural similarity will be scored 10 if the tables share the same structure, including columns, rows, and alignment. The output will be a JSON object with the calculated scores.}
    \label{fig:pscore}
\end{figure*}

For instance, in figure \ref{fig:pscore}, both tables have identical structures, so their structural similarity score is 10. The contents of the first table of Table1 and Table2 are the same, and the second table of Table1 and Table2 are almost 10\% similar. Therefore their content similarity score is 5.

\subsection{H-Score}
\label{H-score}

We attach our algorithm to calculate H-Score as Algorithm \ref{H-score pseudo-code}.
\paragraph{\LaTeX{}}
We use the \href{https://github.com/phfaist/pylatexenc}{\texttt{pylatexenc}} library to parse a given \LaTeX{} table, and walk through the parse-tree structure in the \texttt{tabular} environment to identify the table ``cells''.
We score the content similarity based on strings within the cells, and score structural similarity based on having the matching number of rows and columns, the same caption, and the same cell alignment.

\paragraph{HTML}
We use the \href{https://pypi.org/project/beautifulsoup4/}{\texttt{beautifulsoup4}} library to parse a given \LaTeX{} HTML snippet and walk through the parse-tree structure in \texttt{<table>}, \texttt{<ul>} or \texttt{<ol>} tags to identify data cells. 
We separately build a tree of white-listed HTML tags to score the structural similarity, traversing an HTML document tree structure, disregarding the actual content within the tags and simplifying it by focusing only on specific HTML tags (defined in RECOGNIZED_HTML_TAGS).
We score the content similarity based on strings within the cells and score structural similarity based on the similarity of the structure tree and the total number of cells matching.


White-listed HTML tags:
\label{HTML}

\begin{lstlisting}
 RECOGNIZED_HTML_TAGS = [
    "table", "tr", "th", "td",
    "ul", "ol", "li",
    "div", "span", "p",
    "a", "img", "embed", "pre",
    "h1", "h2", "h3", "h4", "h5", "h6",
    "input", "button",
]
\end{lstlisting}

\paragraph{Raw Text Tables}
In our evaluated dataset, each example consists of two tables (Team and Player). 
We do a string search for \texttt{"Team"} and \texttt{"Player"} headers to identify the two tables.
We then parse the tables according to Markdown formatting, with newlines and pipes as row and column dividers respectively, to identify the table cells.
We score the content similarity based on strings within the cells, and score structural similarity based on the similarity of column names and the number of rows and columns matching.

\paragraph{String Similarity Measurement:} Our script includes methods to calculate the similarity between two strings. These methods can be used to compare the structure or content of HTML, latex documents, or any other pair of strings. The similarity is evaluated using well-established algorithms in text analysis: the Levenshtein distance and the SequenceMatcher from Python's difflib module.

\clearpage

\clearpage

\begin{algorithm*}
\SetAlgoLined
\RestyleAlgo{ruled}
\DontPrintSemicolon
\caption{The provided algorithm is the H-score algorithm with pseudocode. It aims to calculate the content similarity (Content H-score) and format similarity (Format H-score) between predictions and references. The algorithm begins by defining functions for Levenshtein similarity and Difflib similarity to compute string similarities. It then iterates over each pair of predictions and references, parsing their structures and data to determine the similarity in terms of the number of columns, rows, column names, and data rows. The final step involves averaging the similarity scores to obtain the Content H-score and Format H-score.}\label{H-score pseudo-code}
\KwData{predictions $\alpha$, references $\beta$}
\KwResult{Content H-score $\gamma$, Format H-score $\delta$}
\SetKwFunction{FLevSim}{LevenshteinSimilarity}
\SetKwProg{Fn}{Function}{:}{}
\Fn{\FLevSim{\textit{p}, \textit{q}}}{
    \If{\textit{q} is empty}{
        \KwRet 0.0\;
    }
    \textit{s} $\gets$ 1 - LevenshteinDistance(\textit{p}, \textit{q}) / \textit{(2 * length(p)})\;
    \KwRet \textit{s}\;
}
\SetKwFunction{FLevSim}{DifflibSimilarity}
\SetKwProg{Fn}{Function}{:}{}
\Fn{\FLevSim{\textit{p}, \textit{q}}}{
    \textit{s} $\gets$ difflib.SequenceMatcher(\textit{None}, \textit{p}, \textit{q})\;
    \KwRet \textit{s}\;
}
$\gamma$ $\gets$ 0\;
$\delta$ $\gets$ 0\;
\For{$\sigma, \phi$ in $zip(\alpha, \beta)$}{
        Parse $\sigma$ to get its number of columns $\lambda_c$, number of rows $\lambda_r$, column names $\mu$ and data rows $\kappa$\;
        Parse $\phi$ to get its number of columns $\theta_c$, number of rows $\theta_r$, column names $\tau$ and data rows $\omega$\;
        $\rho_1$ $\gets$ \textbf{if} $\lambda_c$ == $\theta_c$ \textbf{then} 1.0 \textbf{else} 0.0 \;\tcp{Whether their number of columns are same}
        $\rho_2$ $\gets$ \textbf{if} $\lambda_r$ == $\theta_r$ \textbf{then} 1.0 \textbf{else} 0.0\;\tcp{Whether their number of rows are same}
        $\rho_3$ $\gets$ \textbf{LevenshteinSimilarity}($\mu, \tau$)\;\tcp{Compute their columns Levenshtein Similarity}
        $\rho_4$ $\gets$ \textbf{DifflibSimilarity}($\mu, \tau$)\;\tcp{Compute their columns Difflib Similarity}
        $\rho_5$ $\gets$ \textbf{LevenshteinSimilarity}($\kappa, \omega$)\;\tcp{Compute their data Levenshtein Similarity}
        $\rho_6$ $\gets$ \textbf{DifflibSimilarity}($\kappa, \omega$)\;\tcp{Compute their data Difflib Similarity}

        $\gamma \gets \gamma + average(\rho_5, \rho_6)$\;
        $\delta \gets \delta + average(\rho_1, \rho_2, \rho_3, \rho_4)$\;
    }
$\gamma$ $\gets$ $\gamma$ / \textit{length}($\alpha$)\;
$\delta$ $\gets$ $\delta$ / \textit{length}($\alpha$)\;
\end{algorithm*}

\clearpage





\clearpage

\section{Prompt for FormatCoT and Inference}
\label{PromptDescription}

\subsection{Prompt for FormatCoT}
\label{FormatCoTPrompt}
\paragraph{Raw Text Table Description} Traditional data-to-text datasets only have raw text for each table. However, it is not enough for GPT-3.5 or other LLMs to generate correct tables. As a result, we added some format descriptions to help them generate the correct tables. We use GPT-3.5 to achieve this. We want to get detailed format information without concrete contents in cells, so we explicitly include these requirements in the prompt. Here is our prompt: ``Describe details about the given text. First, give the number of tables, and then for each table, describe its format such as the number of columns and rows, column names, and row names.''

\paragraph{HTML Table Description} Unlike data-to-text datasets, HTML datasets only have final outputs, so we are required to generate a detailed description of their format and content. For content descriptions, we can simply ask GPT-3.5 to output raw text without HTML tags. For format descriptions, however, we need to ask GPT-3.5 to describe each tag, otherwise, it will leave out some tags and describe the table in general rather than detailed information. Moreover, it is necessary to ask it to use specific numbers instead of `several' or `multiple'. Here is our prompt for HTML format descriptions: ``Describe the format of this HTML in detail according to each HTML tag of the following HTML code. Be careful and make sure don't miss any HTML tags. Please use more than 300 words to explain the format. Use specific numbers rather than being vague about several.''

\paragraph{LaTeX Table Description}  Similar to HTML prompt generation, it is necessary to ask GPT-3.5 to generate both format descriptions and content descriptions as latex datasets only have final outputs. For content descriptions, we can simply ask GPT-3.5 to describe the given latex table as detailed as it can and include all cells. For format description, since the latex format is too complex, we need to give it a small example to learn. Then we ask GPT-3.5 to describe the detailed format of a given latex table, including specific questions to help it generate format descriptions. Here is our prompt for latex format descriptions:  ``Describe the detailed format of a given latex table according to the commands and tags with more than 500 words. Include: Whether there is table border lines? How is text alignment? What are table attributes? Whether to bold? Whether to add \textbackslash ref? Please clearly explain whether there are horizontal and vertical lines bordering each row and column. Say anything about a special "\textbackslash" format token in latex if there is one. Don't display latex code directly. Use natural language. And provide enough format information for me to recreate this table based on your output description.''

\subsection{Prompt for Inference}
\label{InferencePrompt}
When inferencing raw text tables, LLMs tend to output tabular results rather than raw text tables. As a result, we need to give it an example output first, then tell the model that the input consists of two parts, text and format descriptions, and ask the model to generate the output based on them. For HTML and Latex inference, we can simply ask models to infer from the input and specify the format and content sections in the input, since models can generate correct syntax.

\clearpage

\begin{table*}[h!]
  \centering
  \begin{tabular}{|m{6cm}<{\centering}|m{8cm}<{\centering}|} 
    \hline
    \textsc{FormatCoT} Prompt for Raw Text Table & \vspace{5pt}Describe details about the given text. First, give the number of tables, and then for each table, describe its format such as the number of columns and rows, column names, and row names.\vspace{5pt} \\ 
    \hline
    \textsc{FormatCoT} Prompt for HTML Table & \vspace{5pt}Describe the format of this HTML in detail according to each HTML tag of the following HTML code. Be careful and make sure don't miss any HTML tags. Please use more than 300 words to explain the format. Use specific numbers rather than being vague about several.\vspace{5pt}\\
    \hline
    \textsc{FormatCoT} Prompt for LaTeX Table & \vspace{5pt}Describe the detailed format of a given latex table according to the commands and tags with more than 500 words. Include: Whether there are table border lines? How is text alignment? What are table attributes? Whether to bold? Whether to add \textbackslash ref? Please clearly explain whether there are horizontal and vertical lines bordering each row and column. Say anything about a special "\textbackslash" format token in latex if there is one. Don't display latex code directly. Use natural language. And provide enough format information for me to recreate this table based on your output description.\vspace{5pt}\\ 
    \hline
    Prompt for Inference & \vspace{5pt}Based on the example output above, generate the raw text/HTML/LaTeX  table according to the following description. \vspace{5pt}\\
    \hline
  \end{tabular}
  \caption{Our prompts for \textsc{FormatCoT} and Inference. The prompt requests an overview of table formatting in raw text, HTML, and LaTeX formats, including descriptions of the number of tables, column and row structures, formatting elements, and specific instructions. The prompt for inference illustrates the request to generate the tables based on the provided details.}
  \label{Prompts}
\end{table*}

\clearpage












\clearpage

\section{Ability Map}
\label{map}

Based on our automated evaluation, we selected Vicuna, GPT-3.5, GPT-4, and Ours as representative models and conducted an in-depth analysis of the causes of model errors.

We identified content accuracy, formatting, numerical reasoning, and handling of long tables as the main sources of these errors.

At the fundamental level, we decompose the process of model-generated complex structured outputs into two parts: Content Selection and Format Planning. Initially, the model needs to identify key information from a given vast amount of unstructured input, extract this information, understand it, and organize it. Subsequently, it needs to plan how to summarize these extracted details, devise the format of the table to be generated, and then fill in the information.

Accordingly, we can break down the model's capabilities into Coverage, Formatting Reasoning, Comprehension, Pragmatics, and Hallucination Control.

Coverage entails the model's ability to accurately cover the content in the input. Formatting Reasoning pertains to judgment about the output format, assessing if the model can find the most appropriate and reasonable structured format.

Comprehension reflects whether the model can understand the content of the input, as there are times when it is necessary to infer from a large amount of data (including performing addition or subtraction or comparing multiple elements).

Pragmatics involves the ability to utilize special formats, such as HTML tags and specific syntax in LaTeX.

Finally, Hallucination Control signifies the model's ability to refrain from generating content not present in the input.

We carried out manual annotations and obtained visualized results to demonstrate these aspects.

\clearpage

\section{Ablation Study for FormatCoT}
\label{ablation}
\subsection{Contrast between descriptions}
\begin{figure*}[!ht]
    \centering
    \includegraphics[width=0.65\textwidth]{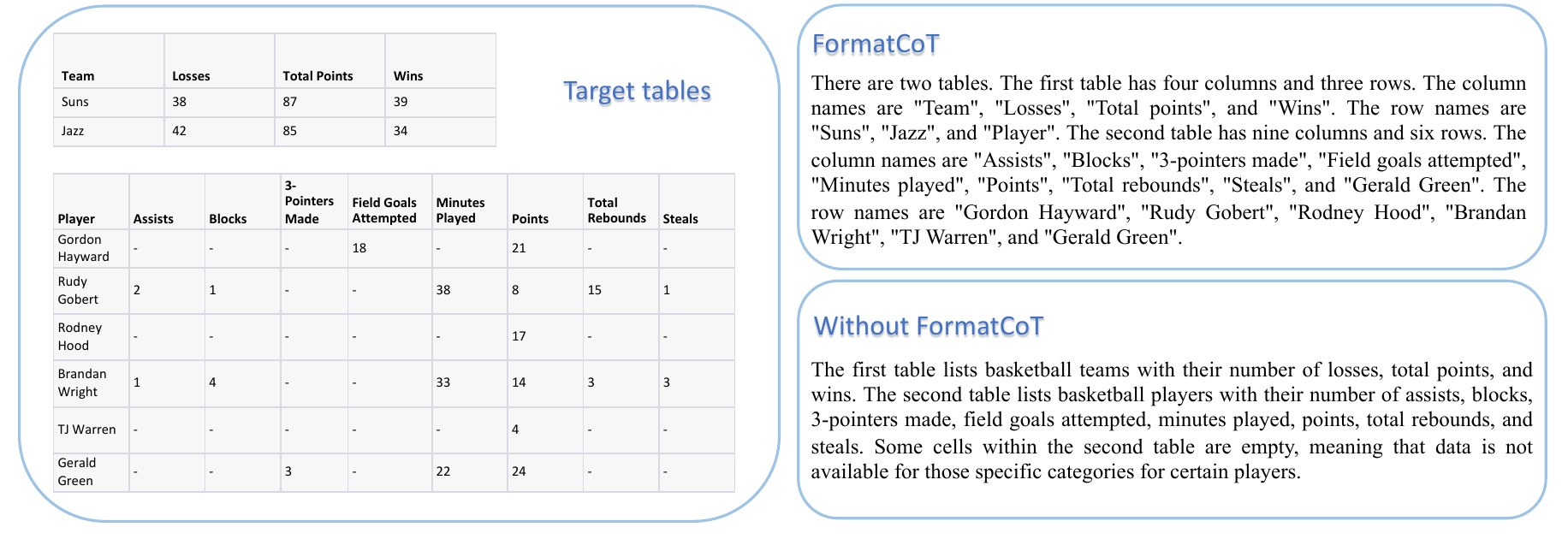}
    \caption{Using \textsc{FormatCoT} and normal instructions to ask GPT-3.5 to generate table descriptions based on the input text, \textsc{FormatCoT} results contain more detailed information about row names.}
    \label{fig:FormatCoT1}
\end{figure*}

In this section, we conduct an ablation study to examine the impact of our proposed \textsc{FormatCoT}. In the generation of table descriptions sans \textsc{FormatCoT}, we simply utilize the prompt: ``Provide a description of the following tables.'' We display the results in figure \ref{fig:FormatCoT1}. The primary differentiation between results pivots on the extent of details incorporated.

For instance, in the \textsc{FormatCoT} result, the description comprises an array of detailed format information - encompassing row names, column names, and table count. The precision in these details proves substantial enough for models to accurately recreate the tables in question.

Contrastingly, the outcome bereft of \textsc{FormatCoT} conveys considerably less information - providing incomplete column names without the accompaniment of row names. This sparse degree of detail proves insufficient for models seeking to faithfully regenerate the corresponding tables.

\subsection{Contrast between results}
\begin{figure*}[!hb]
    \centering
    \includegraphics[width=0.65\textwidth]{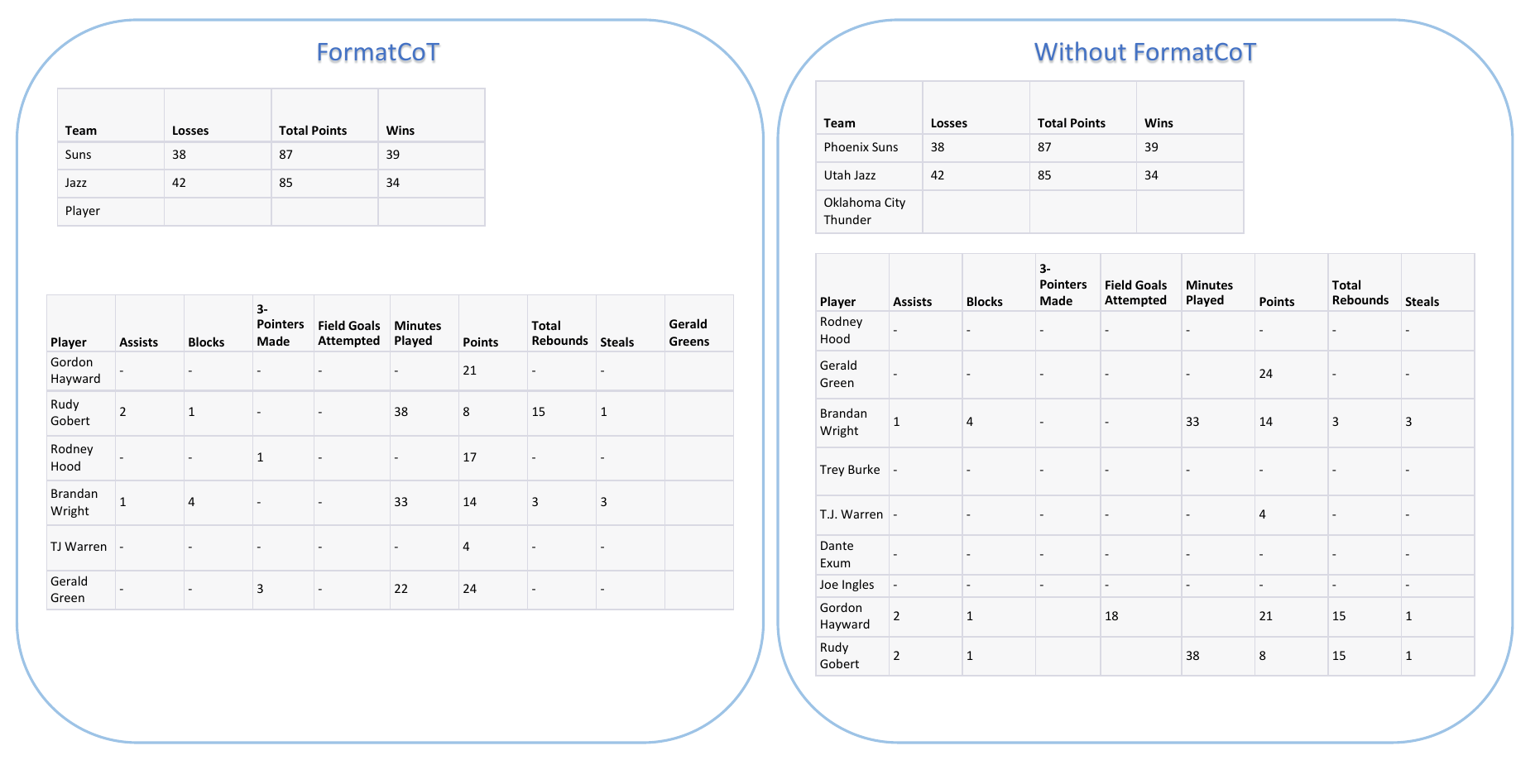}
    \caption{Using two descriptions to regenerate table descriptions based on the input text and descriptions, \textsc{FormatCoT} result is more correct in both format and content.}
    \label{fig:FormatCoT2}
\end{figure*}

In this section, we draw a comparison between two sets of description results, shown in figure \ref{fig:FormatCoT2}. The \textsc{FormatCoT} result showcases a table that stands remarkably close to the correct table, albeit with minor errors. It contains an extra row termed ``Player'' in the initial table, a discrepancy potentially attributable to the fact that the result comprises two tables, with ``Player'' denoting the header of the subsequent table. We posit that this error could potentially be circumvented with a different method of integrating table names.

Furthermore, an additional column surfaces in the second table, which in reality represents the final row of that table. Besides these minor inaccuracies, the \textsc{FormatCoT} result accurately replicates the content in each cell as well as maintaining the overall format.

Conversely, the alternative result contains multiple errors that span both content and format. Initially, an additional row is present in the first table, introducing an unrelated basketball team that bears no relevance to the game under consideration. Following this, the second table possesses an excessive number of player names, encompassing unnecessary players along with coaches who did not participate in the game.

Furthermore, its content is not entirely accurate, with discrepancies present in the statistics attributed to both Gordon Hayward and Gerald Green. These shortcomings underscore the efficiency and essentiality of implementing the \textsc{FormatCoT} to ensure accuracy and precision.

\clearpage

\end{document}